\documentclass[journal]{IEEEtran}
\ifCLASSINFOpdf
\else
\fi

\usepackage{bbding}
\usepackage{eucal}
\usepackage{bbm}

\AtBeginDocument{%
	\providecommand\BibTeX{{%
			\normalfont B\kern-0.5em{\scshape i\kern-0.25em b}\kern-0.8em\TeX}}}

\usepackage{graphicx}
\usepackage{amsmath}

\usepackage{amssymb}
\usepackage{amsthm}
\usepackage{tabularx}
\usepackage{threeparttable}
\usepackage{array}

\usepackage{booktabs} 
\usepackage{color}
\usepackage{algorithm}
\usepackage{algorithmic}
\usepackage[caption=false]{subfig}
\usepackage{multirow}
\usepackage{cite}
\newcommand{\MyMapTemplatePrefix}[4]{\expandafter#1\csname#3#4\endcsname{#2{#4}}}
\newcommand{\MyMapTemplatePrefixNew}[5]{\expandafter#1\csname#4#5\endcsname{#2{#3{#5}}}}

\begin{document}
\bstctlcite{IEEEexample:BSTcontrol}
%
\title{Graph Jigsaw Learning for Cartoon Face Recognition}
%
%
%

\author{Yong~Li,
        Lingjie Lao,
        Zhen~Cui, 
        Shiguang~Shan,
        Jian Yang 
\thanks{Corresponding author: Zhen Cui, zhen.cui@njust.edu.cn}
\thanks{Yong Li, Lingjie Lao, Zhen Cui, Jian Yang are with the Key Laboratory of Intelligent Perception and Systems for High-Dimensional Information, Ministry of Education, School of Computer Science and Engineering, Nanjing University of Science and Technology, Nanjing, 210094, China (e-mail: (yong.li, laolingjie, zhen.cui)@njust.edu.cn, csjyang@mail.njust.edu.cn). }
\thanks{S. Shan is with the Key Laboratory of Intelligent Information Processing of Chinese Academy of Sciences, Institute of Computing Technology, CAS, Beijing 100190, China, and with the University of Chinese Academy of Sciences, Beijing 100049, China, and also with CAS Center for Excellence in Brain Science and Intelligence Technology (e-mail: sgshan@ict.ac.cn).}
\thanks{Manuscript received April 19, 2005; revised August 26, 2015.}}

\markboth{Journal of \LaTeX\ Class Files,~Vol.~14, No.~8, August~2015}%
{Shell \MakeLowercase{\textit{et al.}}: Bare Demo of IEEEtran.cls for IEEE Journals}

\maketitle

\begin{abstract}

Cartoon face recognition is challenging as they typically have smooth color regions and emphasized edges, the key to recognize cartoon faces is to precisely perceive their sparse and critical shape patterns. However, it is quite difficult to learn a shape-oriented representation for cartoon face recognition with convolutional neural networks (CNNs).  To mitigate this issue,  we propose the GraphJigsaw that constructs jigsaw puzzles at various stages in the classification network and solves the puzzles with the graph convolutional network (GCN) in a progressive manner. Solving the puzzles requires the model to spot the shape patterns of the cartoon faces as the texture information is quite limited.  The key idea of GraphJigsaw is constructing a jigsaw puzzle by randomly shuffling the intermediate convolutional feature maps in the spatial dimension and exploiting the GCN to reason and recover the correct layout of the jigsaw fragments in a self-supervised manner.  The proposed GraphJigsaw avoids training the classification model with the deconstructed images that would introduce noisy patterns and are harmful for the final classification.  Specially, GraphJigsaw can be incorporated at various stages in a top-down manner within the classification model, which facilitates propagating the learned shape patterns gradually. GraphJigsaw does not rely on any extra manual annotation during the training process and incorporates no extra computation burden at inference time. Both quantitative and qualitative experimental results have verified the feasibility of our proposed GraphJigsaw, which consistently outperforms other face recognition or jigsaw-based methods on two popular cartoon face datasets with considerable improvements.
\end{abstract}


\IEEEpeerreviewmaketitle

\section{Introduction}

\IEEEPARstart{C}{artoon} face is a pivotal part to understand and interact with the virtual world \cite{zhang2014data, chen2002pictoon, li2011guided}.  Precisely recognizing these cartoon characters is a critical prerequisite for many commercial applications, such as automatic cartoon face verification \cite{zheng2020cartoon}, computer-aided cartoon face generation \cite{zhou20163d}, cartoon movie recommendation \cite{lu2011personalization}.  
With the advent of large-scale face recognition (FR) datasets such as MS-Celeb-1M \cite{guo2016ms}, VGGFace2 \cite{cao2018vggface2}, MegaFace \cite{kemelmacher2016megaface}, WebFace260M\cite{zhu2021webface260m}, the deep-learning-based models have achieved promising recognition and verification accuracies on human faces \cite{wu2018light, schroff2015facenet, huang2020curricularface, liu2017sphereface, wang2018cosface, deng2019arcface}. While human facial recognition technology is becoming increasingly sophisticated and mature, cartoon character recognition is still in its infancy. The performance gap is mainly achieved by the utilization of tremendous manual labeling datasets, which is extremely deficient in cartoon media. 

To fill the gap between the FR accuracy on the human faces and the cartoon faces, researchers make efforts on collecting large-scale cartoon face datasets such as iCartoonFace \cite{zheng2020cartoon}, Danbooru \cite{Anonymous_Danbooru2018}. However, cartoon face recognition is still challenging due to its intrinsic characteristics \cite{wang2020learning, chen2018cartoongan, rios2021daf}, i.e., (1) cartoon faces usually have smooth and sparse color regions, the texture information is quite limited, (2)  details of cartoon faces are mainly outlined by its structure/shape that consists of sharp and clear edges. Thus the key to distinguish different cartoon characters is to precisely spot the sparse and critical shape patterns \cite{rios2021daf}. In biological vision, shape is arguably the most important cue for recognition \cite{baker2018deep}.
However, standard convolutional neural networks (CNNs) actually exhibit a ``texture-bias'' \cite{geirhos2018imagenet, Islam2021_Shape}, they lack the shape representations and processing capabilities that form the primary bases of human object classification \cite{baker2018deep}.

Inspired by the works \cite{chen2019destruction, du2020fine} that spot the fine-grained details in the input images via training with the jigsaw patches, we challenge the problem of cartoon face recognition by introducing an effective jigsaw solving approach to learn the critical shape patterns that are discriminative for cartoon face recognition. 
As texture information is quite limited in the cartoon faces, the classification network has to pay more attention to the local critical shape patterns to solve the jigsaw. 
Specially, to utilize those implicit relations between jigsaw fragments, we build the graph model to describe the topological layout of permutated jigsaws, where each fragment is treated as a graph node in the jigsaw puzzle. To further reason and recover the correct layout of the jigsaw patches, we introduce graph convolutional network (GCN) \cite{kipf2016semi, velivckovic2017graph}  by propagating neighbor information in an iterative manner. As the perception and aggregation of contextual information, the recovery of disordered jigsaws becomes more accessible in a self-supervised mode. Thus, the graph jigsaw model can learn to understand the local to holistic shape information, capture and encode the discriminative shape patterns, so as to identify each patch accurately to infer their affinities and relative placements.
\begin{figure}[htb]
	\centering
	\includegraphics[width=0.48\textwidth]{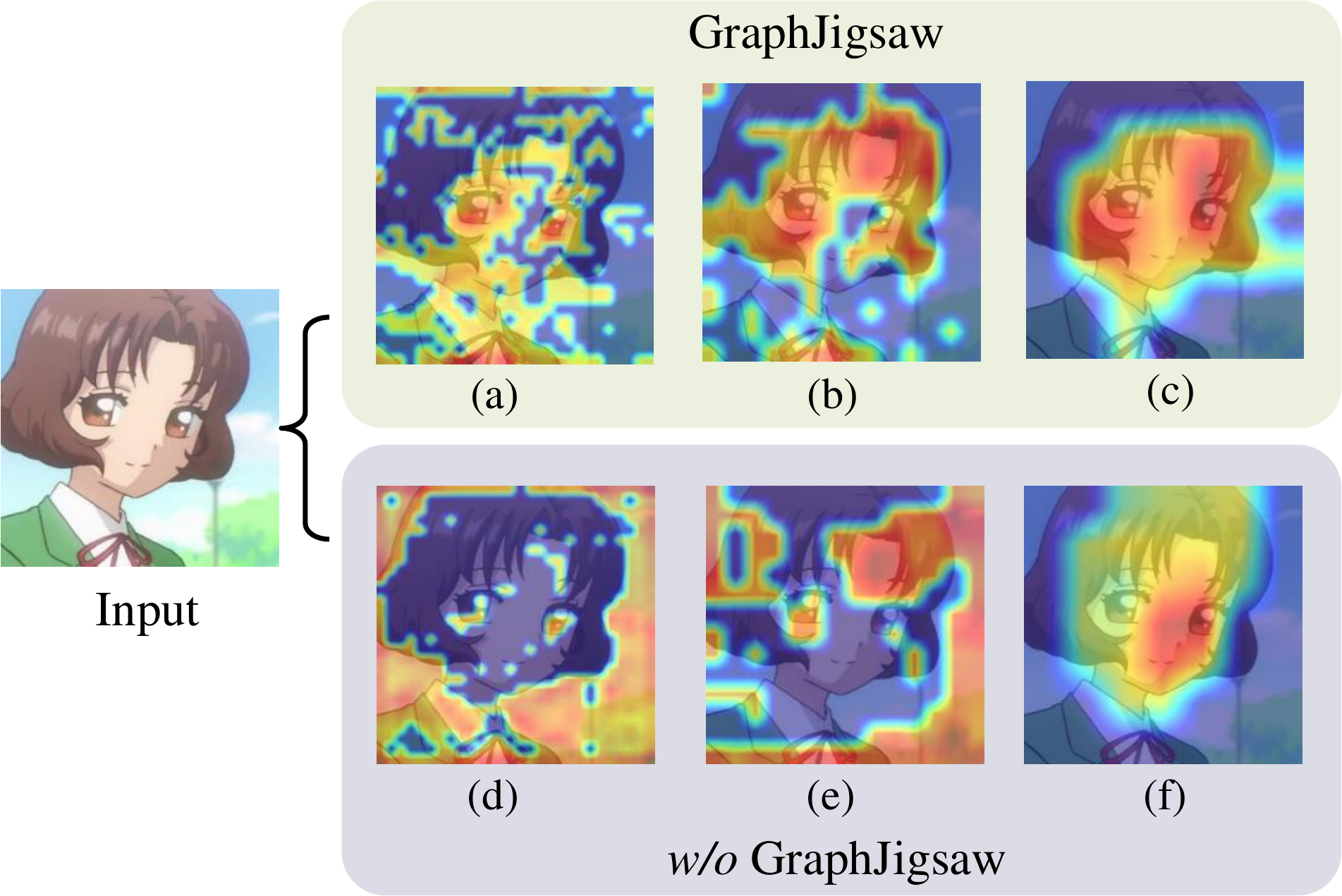}
	\caption{Attention maps of the last three stages’ convolution layers in a ResNet-50 classification model. (a), (b), (c) in the top row show the model with \textit{GraphJigsaw} learns to focus on the most discriminative regions in a progressive manner. (d), (e) in the bottom row illustrate the model without GraphJigsaw fails to perceive the shape characteristics in the early stages and only shows part correct attention in the last stage in (f). Better viewed in color and zoom in.}
	\label{fig:cartoon_fig1}
\end{figure}

It is worth noting that directly training the classification model with images that consist of shuffled patches would introduce inevitable noisy visual patterns that are harmful for the final classification \cite{chen2019destruction}. To tackle this issue, we propose to introduce the jigsaw tasks on the intermediate stages in the classification network. A classification network such as ResNet-50 \cite{he2016deep} typically has various stages where each stage consists of several groups of convolutional blocks, we thus construct one jigsaw puzzle by shuffling the input of each stage in the spatial dimension and then build one graph in the space of the shuffled features. 
To solve the jigsaw puzzle, we exploit the GCN to encode the adjacent correlations among the graph vertices by aggregating their neighbors iteratively. During this encoding process, GraphJigsaw captures  the layout of current disordered states and perceives the characteristics of the input graph. 
We then recover the correct layout of the permuted vertices with the encoded vertices in an adaptative decoding process.
To force the model solving the jigsaw via spotting the shape information rather than the low-level statistics (e.g., the pixel intensity, color distribution, chromatic aberration) \cite{noroozi2016unsupervised}, we propose to self-supervise the GraphJigsaw by the output of each stage in the classification network.
We call the proposed method GraphJigasw as we recover the correct layout of the jigsaw fragments in an adaptive graph encoding and decoding process. 
Our experiments show that this formalization of a jigsaw solver provides a much stronger model for cartoon face recognition. 

Generally, CNN tends to capture the low-level features such as edges, corners, color conjunctions in the early layers \cite{zeiler2014visualizing, liu2017richer}. As the shape patterns are quite sparse in the cartoon faces, they are difficult to be perceived with the standard CNN. As illustrated in the bottom row in Fig.~\ref{fig:cartoon_fig1},  the standard CNN without GraphJigsaw fails to perceive the shape characteristics in the early stages (sub-figures (d), (e)) and only shows part correct attention in the last stage (sub-figure (f)). 
To fully exploit the rich hierarchical information in the CNN and propogate the learned shape patterns gradually, we propose to incorporate the GraphJigsaw in various stages of the classification network in a top-down manner.
As illustrated in the top row in Fig.~\ref{fig:cartoon_fig1}, the model with GraphJigsaw at different stages is capable of focusing on the most discriminative regions in a progressive manner (sub-figures (a-c)). 
With the propressive GraphJigsaw, the valuable shape characteristics of the input cartoon faces can be reserved in the early stages and strengthened in the later stages gradually.  More visual examples can be found in Fig.\ref{fig:cartoon_visualization} in section \ref{sec:ablation_study}.

In summary, the contributions of this work are summarized as follows: 
\begin{itemize}
	\item We propose the GraphJigsaw that incorporates the jigsaw puzzle in the classification network and solves the jigsaw with graph convolutional network.  GraphJigsaw learns to understand the local to holistic shape information in the input cartoon faces in the jigsaw  solving process. GraphJigsaw is only used in the training phase and brings no computational cost during the deployment. 
	\item GraphJigsaw does not rely on extra manual annotation and can be incorporated in various stages in the classification network, which facilitates propagating the learned shape patterns of the input cartoon faces gradually. To our best knowledge, GraphJigsaw is the first work that constructs and solves the jigsaw puzzle on the imtermediate convolutional feature maps.
	\item Extensive experiments on two publicly largest cartoon face datasets verify the superiority of our proposed GraphJigsaw. Our proposed method can serve as a strong backbone to facilitate future research on cartoon face recognition.
\end{itemize}


\section{Related Work}
\label{sec:related_work}
We review the previous work considering two aspects that are related to ours, i.e., the similar tasks (cartoon face recognition) and related techniques (jigsaw puzzles solving).

\textbf{Cartoon Face Recognition.} 
Over the recent years, many literatures aim to automatically generate a cartoon-style face with an input RGB image \cite{chen2018cartoongan, wang2020learning} or sketches \cite{wang2011sketch2cartoon, ci2018user}, while cartoon face recognition has been less explored and remains a challenging problem.
Automatic cartoon face recognition performs poorly in daily life applications due to the lack of large-scale real-world datasets. Existing datasets (WebCaricature \cite{huo2017webcaricature}, IIIT-CFW \cite{mishra2016iiit}, Manga109 \cite{fujimoto2016manga109}) contain limited examples and fail to reflect the true distribution of the cartoon faces.
Among the existing methods, Takayama et al \cite{takayama2012face} extracted features that reflect skin color, hair color, hair quantity for cartoon face recognition.
 Saito et al \cite{saito2015illustration2vec} proposed a CNN-based model for cartoon image retrieval. 
Zhou et al. \cite{zhou2018toonnet} constructed a ToonNet that contains thousands of cartoon-styled images and introduced several techniques for building the deep neural network towards cartoon face recognition.
Zheng et al. \cite{zheng2020learning} proposed a meta-continual learning method that is capable of jointly learning from heterogeneous modalities such as sketch, cartoon, and caricature images. 
These pioneering methods brings inspiration for the research on cartoon face recognition. 
In this work, we study the automatic cartoon face recognition by adopting the iCartoonFace  \cite{zheng2020cartoon} and Danbooru \cite{Anonymous_Danbooru2018} datasets, which are the largest and most comprehensive datasets for cartoon face recognition. We argue that the shape characteristics can be better exploited for accurate cartoon face recognition.

\textbf{Jigsaw Puzzles Solving.} 
 There are many literatures on solving jigsaw puzzles computationally \cite{cho2010probabilistic, pomeranz2011fully, liu2011automated, son2014solving, bridger2020solving} that rely on the low-level texture statistics across fragments. However, solving a Jigsaw puzzle based on these cues does not require to full understand the global shape of the object. 
Besides, jigsaw solving has been studied for pretraining \cite{noroozi2016unsupervised}, fine-grained image recognition \cite{chen2019destruction, du2020fine}, mixed-modal image retrieval \cite{pang2020solving}.
Noroozi et al \cite{noroozi2016unsupervised} posed jigsaw solving as a permutation recognition task and introduced the context-free network (CFN). CFN takes image patches as input to learn both a feature mapping of object parts as well as their correct spatial arrangement. 
Pang et al \cite{pang2020solving} formulated the puzzle in a mixed-modality fashion and learned a pre-training model for fine-grained sketch-based image retrieval.
For fine-grained recognition, Chen et al \cite{chen2019destruction} proposed a novel ``Destruction and Construction Learning'' (DCL) method that enhances the difficulty of fine-grained image recognition by directly training with the destructed images and estimating the locations of the shuffled patches.
Du et al \cite{du2020fine} proposed a random jigsaw patch generator that encourages the network to learn features at specific granularities with a progressive training strategy (PMG).
Compare with DCL and PMG, our proposed GraphJigsaw does not adopt deconstructed images for training and avoids introduce noisy visual patterns. 
We treat each fragment in the jigsaw puzzle as a graph node and exploit GCN to reason the proximity and relative placements of the fragments in the jigsaw puzzle.
By solving the puzzle with the GCN in a self-supervised mode, GraphJigsaw learns to identify the discriminative local to global shape patterns of the input cartoon faces that consist of little texture information.

\section{Graph based Jigsaw puzzle Solver}

 This paper aims to propose a graph convolutional network based jigsaw solver for cartoon face recognition. The proposed GraphJigsaw does not require any external supervision or additional manual labels. By adding  GraphJigsaw into the classification model, i.e., shuffling the preceding block of the network and then solve the jigsaw implicitly by mimicing the spatial layouts of a deeper block, the network can learn to strengthen its representations and fully perceive the structural characteristics of the input cartoon faces.

\begin{figure*}
	\centering
	\includegraphics[width=1.0\textwidth]{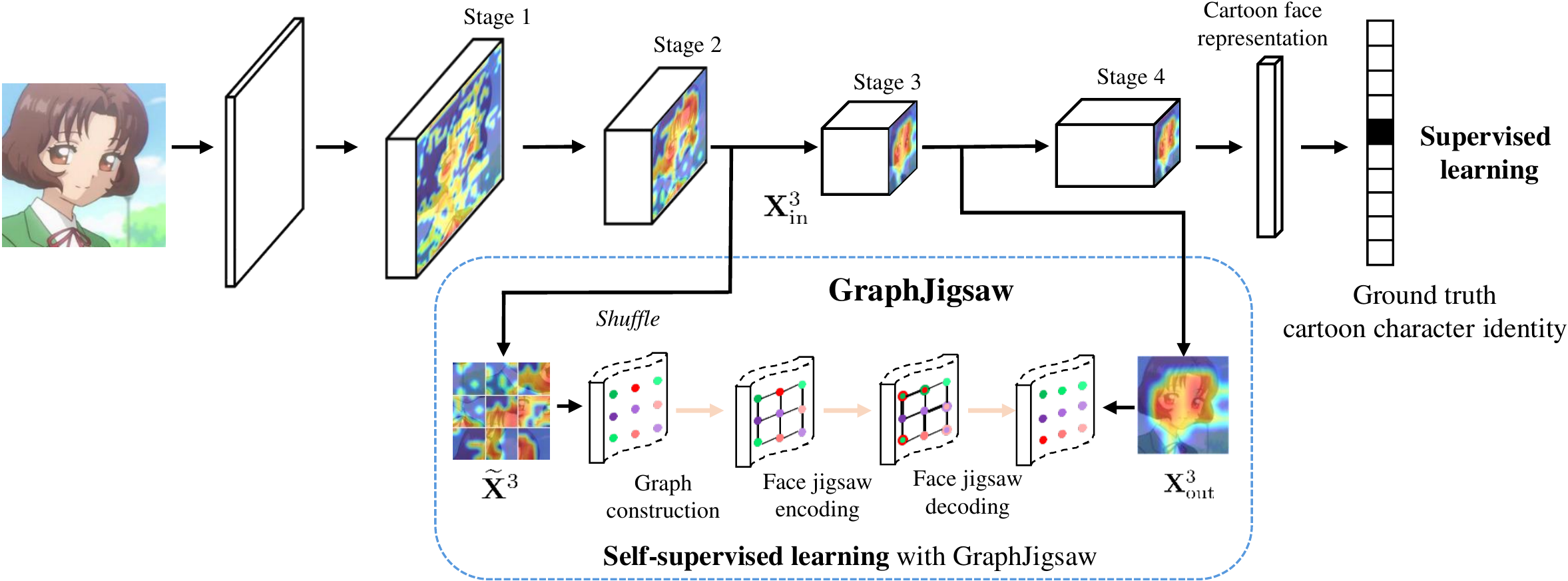}
	\caption{Illustration of the proposed GraphJigsaw. 
		$\mathbf{X}^3_{\text{in}}$ and $\mathbf{X}^3_{\text{out}}$ denote the input and output of the third stage in the backbone network.
		 GraphJigsaw takes $\mathbf{X}^3_{\text{in}}$ as input and constructs a shuffled graph, then GraphJigsaw solves the jigsaw puzzle in the face graph encoding and decoding process.
		 GraphJigsaw can be incorporated in each stage in the backbone network in a top-down manner, thus the valuable shape patterns of the input cartoon faces can be reserved in the early stages and strengthened in the later stages gradually. 
	}
	\label{fig:cartoon_fig2}
\end{figure*}

Fig. \ref{fig:cartoon_fig2} illustrates the framework of our proposed GraphJigsaw, which consists of three parts: (1) Graph construction. (2) Face jigsaw encoding. (3) Face jigsaw decoding.
In graph construction, GraphJigsaw shuffles the input feature map in the spatial dimension and constructs a connected graph $\mathcal{G}$ with the shuffled fragments.
In jigsaw encoding, GraphJigsaw encodes and perceives the adjacent correlations among the input graph vertices.
In jigsaw decoding, GraphJigsaw decodes the input graph and recovers the correct layout of the shuffled patches.  Below, we present the details of the three parts in GraphJigsaw.

\subsection{Shuffled Graph}

Our proposed GraphJigsaw can be incorporated into those conventional convolution networks such as ResNet \cite{he2016deep}.
A ResNet typically consists of several stages (or block groups), among them $\{c2, c3, c4, c5\}$ are commonly used.
The input feature map at the $s$-th intermediate stage is denoted as $\mathbf{X}^s_{in}\in\mathbb{R}^{C_s \times H_s \times W_s}$, where $C_s$, $H_s$, $W_s$ are the number of channels, height, width of the feature map. For convenience, we reset the initial value of $s$ as 1, i.e., $s = \{1, 2, 3, 4\}$ w.r.t $\{c2, c3, c4, c5\}$.
Similarly, we denote the output feature map at the $s$-th stage with $\mathbf{X}^s_{out}\in\mathbb{R}^{C_s' \times H_s' \times W_s'}$.
Here, our objective is to incorporate the proposed GraphJigsaw at different intermediate stages, so that those subtle stripes of cartoon faces can be captured in a self-supervised mode.

To incorporate the jigsaw task, we randomly shuffle $X^s_{in}$ in the spatial dimension to obtain one jigsaw puzzle, formally,
\begin{align}
	\widetilde{X}^s = shuffle(X^s_{in}), \label{equ:shuffle}
\end{align}
where $\widetilde{X}^s$ is the shuffled representations, and $shuffle(\cdot)$ is an permutation operation in the 2D plane space. To reduce the computation burden, we may properly downscale the size of the feature map before shuffling, which however cannot significantly degrade the performance in our experience. Taking the classic pooling strategy for example, we can obtain the small-scale feature map,
\begin{align}
	X^s_{in} \leftarrow \mathcal{P} (X^s_{in}), \label{equ:pooling}
\end{align}
where $\mathcal{P}$ means the adaptive 
average pooling operation. Suppose the pooling size is $M\times M$, we can obtain the shuffled feature $\widetilde{X}^s \in \mathbb{R}^{C_s \times M \times M}$ through the calculation on Eqn.~(\ref{equ:pooling}) and Eqn.~(\ref{equ:shuffle}).

After spatial shuffling, the new representation\footnote{For simplification, we omit the state symbol $s$ in the superscript in the following sections.} $\widetilde{X}$ tends to lose the internal spatial topology information. 
To mine and recover the spatial relations, we define one graph $\mathcal{G}$ in the space of the shuffled feature, named shuffled graph. 
Concretely, we treat each spatial position as one node, and the feature vector at one position as the node attribute. 
Formally, the shuffled graph is denoted as $\mathcal{G}=(\mathcal{V},\mathbf{A},\mathbf{Z})$, where $\mathcal{V}=\{v_1,\cdots, v_{M^2}\}$ is the node set with the size $|\mathcal{V}|=M^2$, $\mathbf{A}\in\mathbb{R}^{M^2\times M^2}$ is the adjacency matrix of nodes, and $\mathbf{Z}\in\mathbb{R}^{C_s\times M^2}$ is the attribute matrix. 
The attribute matrix $\mathbf{Z}$ can be obtained by flattening the feature map in the spatial dimension, i.e., $\mathbf{Z}\leftarrow flatten_{2,3}(\widetilde{x}^s)$, where the subscripts denote the operated dimensionality. 
Thus, the feature vector w.r.t the $i$-th node may be denoted with $\widetilde{X}_i$, which corresponds to the $i$-th column of $\mathbf{Z}$. 
For the adjacency matrix $\mathbf{A}$, we directly employ the spatial position relations of nodes in the shuffled space. If vertices $v_i$ and $v_j$ are spatial neighborhood, then $A_{ij}=1$, otherwise $A_{ij}=0$. 
One may raise the problem that the shuffled feature $\widetilde{X}$ has discarded the original spatial order after random permutation. 
Actually, the permutation relations are just used to favor in self-supervision learning, which attempts to recover the original orders. And, the effect of adding edges is to describe/momerize the current topological structures, such that this jigsaw puzzle can be well solved.

\subsection{Face Jigsaw Encoding}
\label{sec:face_jigsaw_encoding}

Given one shuffled graph $\mathcal{G} = (\mathcal{V},\mathbf{A},\mathbf{Z})$, GraphJigsaw needs to encode and perceive the adjacent correlations among the input graph vertices for the sake of better decoding at the next stage. To this end, we use the signal diffusion on graphs. Concretely, the encoding process follows a neighborhood aggregation strategy, where we iteratively update the representation of a vertex by aggregating its neighbors. After $t$ iterations of aggregation, a vertex's representation can capture the structural information of its $t$-hop neighbor region.
For $i$-th node $v_{i}\in\mathcal{V}$, the feature aggregation can be formulated as:
\begin{align}
	\mathbf{z}^{(t)}_{i} = \phi({\textsc{Agg}}\{(\mathbf{z}_j^{(t-1)},A_{ij})|v_j\in\mathcal{N} (v_i)\}), t=1,\cdots, T,\label{equ:node_encoding}
\end{align}
where $\phi$ is one mapping function, $\textsc{Agg}$ means an aggregation operation such as max-pooling or weighting summary, $t$ means the number of iterations, $\mathcal{N}(v)$ is a set of nodes adjacent to $v$, $T$ is the total iteration number. Similar to convolution on graphs, the aggregation process may be designed as one network layer, formally,
\begin{align}
	\mathbf{z}^{(t)}_{i} = \sigma \left(\mathbf{W}^{(t-1)}\sum_{v_j\in\mathcal{N}(v_i)} D_i^{-1/2}A_{ij}D_j^{-1/2}\mathbf{z}_j^{(t-1)}+\mathbf{b}^{(t-1)}\right), \label{equ:node_encoding_gcn}
\end{align}
where $\mathbf{W},\mathbf{b}$ is a learnable projection matrix and bias vector, $D_i$ is the degree of node $v_i$, and $\sigma$ denotes one nonlinear activation function such as ReLU. Each iteration can be viewed as one convolution layer, so increasing iterations can enlarge the perception scope. Jigsaw encoding aggregates messages of the surrounding vertices, thus captures the layout of current disordered states. In other words, the characteristics of the input graph can be well perceived and encoded in such a Jigsaw encoding process.

\subsection{Face Jigsaw Decoding}
\label{sec:face_jigsaw_decoding}

In order to solve the jigsaw puzzle, we need to recover the correct layout of the permuted vertices. To this end, we design a graph decoding module to re-organize the jigsaw. 
The output of Jigsaw encoding, $\breve{\mathbf{Z}} = [\mathbf{z}_1^{(T)}, \mathbf{z}_2^{(T)}, \cdots, \mathbf{z}_{M^2}^{(T)}]\in\mathbb{R}^{d'\times M^2}$ is used as the input of the graph decoding process. To mine the correlation of jigsaw patches in the decoding part, we introduce the self-attention mechanism \cite{velivckovic2017graph} by computing the attention coefficients between each pair of vertices. The advantage is two-fold: i) better support the recovery of spatial positions, and ii) reduce the difficulty of self-supervision learning. Formally, we can define the attention weight between nodes $v_i$ and $v_j$ as
\begin{equation} \label{equ:self_attention}
	\alpha_{ij} = \sigma(mlp(\mathbf{W}^{\text{a}}\breve{\mathbf{x}}_i, \mathbf{W}^{\text{a}}\breve{\mathbf{x}}_j )),
\end{equation}
where $mlp$ means a single-layer feedforward neural network, $\sigma$ denotes the activation function, and $\mathbf{W}^{\text{a}}$ is the learned parameter. 
To make the learned coefficients comparable across different vertices, we normalize the coefficients for each vertice using the softmax function:
\begin{equation}\label{equ:softmax_normalize}
		\alpha_{ij} \leftarrow \frac{exp(\alpha_{ij})}{\sum_{k=1, k<=M}exp(\alpha_{ik})}.
\end{equation}

To recover Jigsaw orders, we take the deconvolutional strategy similar to the encoding process as used in Eqn.~(\ref{equ:node_encoding}) and Eqn.~(\ref{equ:node_encoding_gcn}). 
The attention mechanism is integrated into this process. Actually, when the attention weights are directly used as the adjacency relations in the matrix $\mathbf{A}$, the adjacency matrix adaptively encodes the feature-wise correlation between the vertices. Thus we may integrate the shuffled states into the attention matrix in order to properly enhance the learning ability of self-supervision, i.e., $A_{ij}\leftarrow \alpha_{ij}\times \textsf{1}_{v_i\in\mathcal{N}(v_j)}$, where $\textsf{1}$ is the indicator function. Through the attention adjacency matrix, we can properly diffuse the messages of vertices and decode the jigsaw puzzle after stacking $T$ deconvolutional layers like Eqn.~(\ref{equ:node_encoding_gcn}).

To constrain the decoding process, a natural choice is to supervise GraphJigsaw with the original un-shuffled representation $\mathbf{X}_{in} \in \mathbb{R}^{C_s \times M \times M}$. 
However, the model will exploit low-level statistics, such as the pixel intensity/color distribution, and chromatic aberration to solve the jigsaw puzzle \cite{noroozi2016unsupervised}. 
To avoid this side effect and force the classification model to absorb the desired structure information, we adopt the output feature map for $s$-th stage $\mathbf{X}_{out} \in \mathbb{R}^{C' \times H' \times W'}$ as the self-supervised signal. 
To match the feature map size, we downscale the spatial size of $\mathbf{X}_{out}$ with the similar pooling operation $\mathcal{P}$. 
Therefore, at the $s$-th stage, the discrepancy between the input shuffled information and the corrected layout can be defined as,

\begin{equation}\label{equ:stage_loss}
  \zeta_{\text{jig}}^s = \|Dec(Enc(shuffle(\mathcal{P}(\mathbf{X}_{in}^s)))) - \mathcal{P}(\mathbf{X}_{out}^s)\|_F^2,
\end{equation}
where $Enc$ and $Dec$ denote the graph encoding and graph decoding part, respectively.
In this Jigsaw encoding and decoding process, GraphJigsaw learns to understand the critical shape information of the input cartoon faces, capture and encode the discriminative shape patterns to identify each patch accurately to infer their semantic correlations. Thus the representation learned by the classification model is reinforced and strengthened. We will verify the effectiveness of GraphJigsaw in section \ref{sec:experimental}.

\subsection{Integration with the Classifier} Apart from training cartoon face recognition network with the conventional softmax loss, we incorporate the stage-wise GraphJigsaw in a top-down manner into each stage of the classification network. Let $S$ denotes the total stages, the full objective function can be formulated as,
\begin{equation}
		\zeta = \zeta_{\text{cls}} + \lambda \sum_{s=1}^{S}\zeta_{\text{jig}}^s, \label{equ:total_loss}
\end{equation}
where $\zeta_{\text{cls}}$ is the softmax loss used for cartoon face recognition, $\lambda$ means a trade-off factor that control the importance of the GraphJigsaw constraint. 

\section{Experiment}
\label{sec:experimental}
In this section, we validate the effectiveness of the proposed GraphJigsaw on two recently released cartoon face datasets: iCartoonFace \cite{zheng2020cartoon}, Danbooru \cite{Anonymous_Danbooru2018}. After the implementation details in Section \ref{sec:implementation_details},  we compare GraphJigsaw with the state-of-the-art face recognition methods and the representative jigsaw-based methods. Then, we analyze different configurations of GraphJigsaw with comprehensive ablation studies.
Finally, we visualize the learned attention maps and the image retrieval results  to qualitatively verify the effectiveness of the proposed GraphJigsaw.

\subsection{Implemental details}
\label{sec:implementation_details}

\textbf{Training:} We evaluate GraphJigsaw on two ResNet-based classification networks \cite{he2016deep}, i.e., ResNet-50 and ResNet-101, and a DenseNet-based classification network \cite{huang2017densely}, i.e. DenseNet169.
The input images are resized to a fixed size of $256 \times 256$ and randomly cropped into $224 \times 224$. Random horizontal flip is exploited for data augmentation. 
A ResNet or DenseNet typically consists of 4 stages (or block groups), each stage in a ResNet consists of multiple bottleneck blocks with residual connections.
Similarly, each stage in a DenseNet contains a densely connected block.
For each stage in the classification network, GraphJigsaw takes the input of the stage for one jigsaw puzzle creation and the output of the stage as the self-supervised signal for solving the puzzle. During the training process, the category label of the cartoon face images is the only manual annotation used for training. We implemented all the experiments using PyTorch \cite{paszke2017automatic} on a Titan-X GPU with 12GB memory. During the training stage, we set the actual batch size as 96, 60 and 64 for the ResNet-50, ResNet-101 and DenseNet-169 models, respectively. The loss weight $\lambda$ in Equ.~\ref{equ:total_loss} was set as 0.1 by grid search.
We trained our proposed model for 60 epochs until convergence.
Without special mention, the default size $M \times M$ of the jigsaw puzzle is set to $3 \times 3$ in this paper. The influence of the choice of $M$ is discussed in Section \ref{sec:ablation_study}.

Directly training the whole network with all the GraphJigsaws at all the four stages simultaneously makes the training hard to coverage. In contrast to that, we propose to incorporate a GraphJigsaw in a top-down manner at each training iteration, this allows the model to mine discriminative shape patterns of the input cartoon faces from local details to global structures when the features are gradually sent into higher stages. We verify the effectiveness of this progressive training strategy in Section \ref{sec:ablation_study}.

\textbf{Inference:} At the inference phase, we merely input the original cartoon face images into the trained model and the GraphJigsaw is unnecessary. It means the proposed method does not introduce computational overhead at inference time.

\textbf{Dataset:} We adopt the iCartoonFace \cite{zheng2020cartoon} and  Danbooru \cite{Anonymous_Danbooru2018} dataset for training and evaluation. \textbf{iCartoonFace} is a recently released challenging benchmark dataset  for cartoon face recognition. It consists of 389,678 images of 5,013 cartoon characters annotated with identity, bounding box, pose, and other auxiliary attributes. iCartoonFace is currently the largest-scale, high-quality and rich-annotated dataset in
the field of cartoon face recognition. The test set of iCartoonFace consists of a gallery set and a probe set. The gallery set consists of 2,500 images. The probe set consists of 20,000 images from 2,000 cartoon characters, where the number of images for each character ranges from 5 to 17.
\textbf{Danbooru} is an anime character recognition dataset that includes about 0.6M (0.54M images for training and 10k images for testing) images. We excluded the categories that have less than 10 images in the training set and obtained approximately $47k$ images from 5127 classes for training. The number of the per-category images ranges from 10 to 10289.  The test set consists of 10000 images from 3670 categories. For the test set in Danbooru dataset, we excluded the categories that merely consist one image and obtained 7622 images from 1292 classes for evaluation. The images for each character in the test set ranges from 2 to 199.

\textbf{Evaluation Metric:} Following \cite{zheng2020cartoon}, we present the experimental results of cartoon face recognition with the identification rate of Rank@K. In the identification test setting, the probe set includes $N$ cartoon persons and each cartoon person has $M$ images. We test each of the $M$ images per cartoon person by adding it to the gallery of distractors and use each of the other $M-1$ images as a probe.
Given a probe image, we rank orders of all images in the gallery set based on their similarity to the probe image. Thus Rank@K is obtained by computing the percentage of the $K$-th shot in the sorted similarity list.


\begin{table}[htb]
	\centering
	\caption{Recognition performance of GraphJigsaw, stage-of-the-art face recognition and the jigsaw-based methods on the iCartoonFace dataset. All the methods use the ResNet-based network structures. \textbf{Bold} denotes the best.}
	\label{tab:iqiyi_sate_of_the_art_resnet50}
	\begin{tabular}{c|c|c|c|c}
		\hline
		Method  & Backbone & Rank@1 & Rank@5 & Rank@10  \\
		\hline
		\multirow{2}{*}{SoftMax}  & ResNet-50 &  75.14 &  88.23 & 91.46 \\
		& ResNet-101 &  76.90 &  89.62 & 92.88 \\
		\hline
		\multirow{2}{*}{CosFace\cite{wang2018cosface}}  & ResNet-50  & 76.34 & 86.64 & 89.05 \\
		& ResNet-101 &  78.12 &  88.52 & 90.98 \\
		\hline
		\multirow{2}{*}{ArcFace\cite{deng2019arcface}}  & ResNet-50 & 76.38 & 90.04 & 92.75 \\
		& ResNet-101 &  79.79 &  90.39 & 92.71 \\
		\hline
		\multirow{2}{*}{DCL\cite{chen2019destruction}} & ResNet-50 & 63.31 & 78.88 & 83.68 \\
	    & ResNet-101 & 66.18 & 82.64 & 88.14 \\
		\hline
		\multirow{2}{*}{PMG\cite{du2020fine}} & ResNet-50 & 80.38 & 91.76 & 94.26 \\
		& ResNet-101 & 81.71 & 92.79 & 95.41 \\
		\hline
		\multirow{2}{*}{\textbf{GraphJigsaw}} & ResNet-50 & \textbf{83.65} & \textbf{93.42} & \textbf{95.42} \\
		& ResNet-101 & \textbf{84.63} &  \textbf{93.92} & \textbf{95.83} \\
		\hline
	\end{tabular}
\end{table}

\begin{table}[htb]
	\centering
	\caption{Recognition performance of GraphJigsaw, stage-of-the-art face recognition and the jigsaw-based methods on the iCartoonFace dataset. All the methods use the DenseNet-169 network structure. \textbf{Bold} denotes the best.}
	\label{tab:iqiyi_sate_of_the_art_densenet}
	\begin{tabular}{c|c|c|c|c}
		\hline
		Method   & Backbone & Rank@1 & Rank@5 & Rank@10  \\
		\hline
		{SoftMax} &DenseNet-169 & 74.15 & 87.81 & 91.28 \\
		\hline
		{CosFace\cite{wang2018cosface}}   & DenseNet-169 & 80.19 & 90.45 & 92.69 \\
		\hline
		{ArcFace\cite{deng2019arcface}}  & DenseNet-169 & 81.56 & 91.09 & 93.26 \\
		\hline
		{DCL\cite{chen2019destruction}}  & DenseNet-169 & 68.85 & 83.41 & 88.76 \\
		\hline
		{PMG \cite{du2020fine}}  & DenseNet-169 & 81.27 & 91.89 & 94.98 \\
		\hline
		{MTD\cite{zheng2020cartoon}} & DenseNet-169 & 83.96 & 93.43 & 95.46 \\
		\hline
		\textbf{GraphJigsaw} &DenseNet-169 & \textbf{85.33} & \textbf{94.41} & \textbf{96.22} \\
		\hline
	\end{tabular}
\end{table}

\begin{table}[htb]
	\centering
	\caption{Recognition performance of GraphJigsaw, stage-of-the-art face recognition and jigsaw-based methods on the Danbooru dataset. All the methods use the ResNet-based network structure. \textbf{Bold} denotes the best.}
	\label{tab:danboru_sate_of_the_art_resnet50}
	\begin{tabular}{c|c|c|c|c}
		\hline
		Method   & Backbone & Rank@1 & Rank@5 & Rank@10  \\
		\hline
		\multirow{2}{*}{SoftMax}   & ResNet-50 & 43.50 & 59.39 & 65.25 \\
		& ResNet-101 & 45.58 & 59.69 & 65.33 \\
		\hline
		\multirow{2}{*}{CosFace\cite{wang2018cosface}}   & ResNet-50 & 45.74 & 62.88 & 69.87 \\
		& ResNet-101 & 47.44 & 64.29 & 70.16 \\
		\hline
		\multirow{2}{*}{ArcFace\cite{deng2019arcface}}  & ResNet-50 & 46.54 & 66.83 & \textbf{71.39} \\
		& ResNet-101 & 48.69 & 67.25 & 72.21 \\
		\hline
		\multirow{2}{*}{DCL\cite{chen2019destruction}}  & ResNet-50 & 31.82 & 46.23 & 52.73 \\
		& ResNet-101 & 33.08 & 48.89 & 55.77 \\
		\hline
		\multirow{2}{*}{PMG \cite{du2020fine}}  & ResNet-50 & 48.21 & 60.47 & 64.97 \\
		& ResNet-101 & 49.85 & 63.48 & 68.12 \\
		\hline
		\multirow{2}{*}{\textbf{GraphJigsaw}} &ResNet-50 & \textbf{52.74} & \textbf{67.40} & 71.15 \\
		& ResNet-101 & \textbf{54.17} & \textbf{71.15} & \textbf{76.09} \\
		\hline
	\end{tabular}
\end{table}

\subsection{Comparison with state-of-the-arts}
\label{sec:comparison_stage_of_the_art}

We compare GraphJigsaw with state-of-the-art face recognition (FR) methods and representative jigsaw-based approaches. For FR methods, we exploit the same classification backbone network with three popular FR losses: softmax, CosFace \cite{wang2018cosface}, ArcFace \cite{deng2019arcface}. Among them, 
CosFace reformulates the softmax loss as a cosine loss and introduces a cosine margin term to maximize the decision margin in the angular space.  ArcFace directly optimizes the geodesic distance margin by virtue of the exact correspondence between the angle and arc in the normalized hypersphere to obtain highly discriminative face recognition features.
We additionally compare GraphJigsaw with the method in \cite{zheng2020cartoon} on the iCartoonFace dataset. In \cite{zheng2020cartoon}, zheng et al. jointly utilizes the human and cartoon training images with various discriminative regularizations in a multi-task domain adaptation manner. We term the method in \cite{zheng2020cartoon} as MTD, as illustrated in Tab.~\ref{tab:danbooru_sate_of_the_art_densenet}.

For the jigsaw-based approaches, we compare GrpahJigsaw with DCL \cite{chen2019destruction} and PMG \cite{du2020fine}, the details of the two methods have been introduced in Section \ref{sec:related_work}.

\begin{table}[htb]
	\centering
	\caption{Recognition performance of GraphJigsaw, stage-of-the-art face recognition and the jigsaw-based methods on the Danbooru dataset. All the methods use the DenseNet-169 network structure. \textbf{Bold} denotes the best.}
	\label{tab:danbooru_sate_of_the_art_densenet}
	\begin{tabular}{c|c|c|c|c}
		\hline
		Method   & Backbone & Rank@1 & Rank@5 & Rank@10  \\
		\hline
	    {SoftMax} & DenseNet-169 & 43.76 & 57.78 & 64.00 \\
		\hline
		{CosFace\cite{wang2018cosface}}   & DenseNet-169 & 48.24 & 66.68 & 71.44 \\
		\hline
    	{ArcFace\cite{deng2019arcface}}  & DenseNet-169 & 49.30 & 67.42 & 73.32 \\
		\hline
		{DCL\cite{chen2019destruction}}  & DenseNet-169 & 34.79 & 49.22 & 56.15 \\
		\hline
	    {PMG \cite{du2020fine}}  & DenseNet-169 & 50.60 & 63.44 & 67.07 \\
		\hline
		\textbf{GraphJigsaw} &DenseNet-169 & \textbf{55.60} & \textbf{71.60} & \textbf{77.90} \\
		\hline
	\end{tabular}
\end{table}

\begin{figure*}[htb]
	\centering

 	\subfloat[CMC curves on the \textbf{iCartoonFace} dataset (ResNet-50)]
    {
    \includegraphics[width=0.45\linewidth]{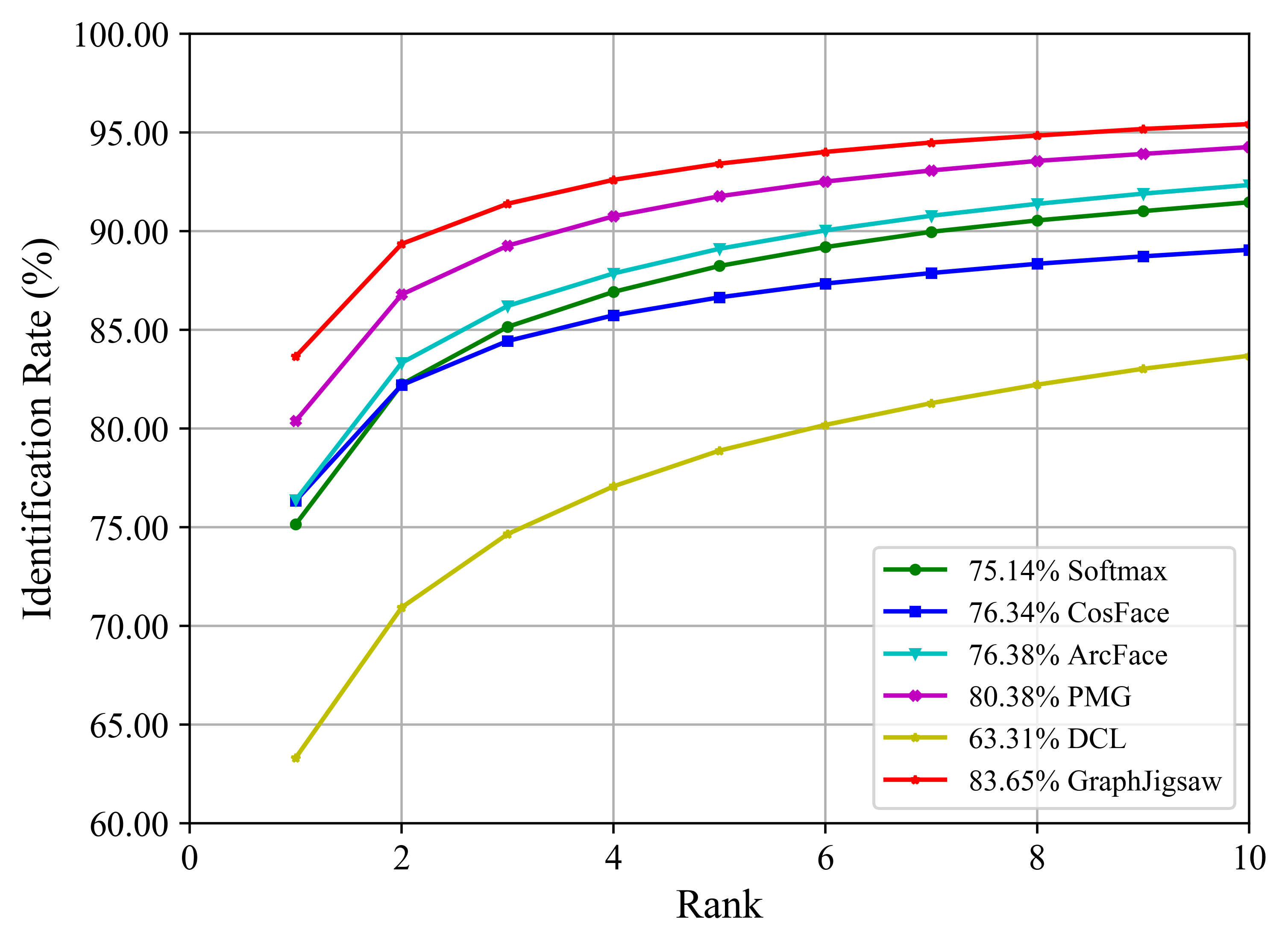}
		\label{fig:resnet-iqiyi}
    }\hfil   
 	\subfloat[CMC curves on the \textbf{iCartoonFace} dataset (DenseNet-169)]
    {
    \includegraphics[width=0.45\linewidth]{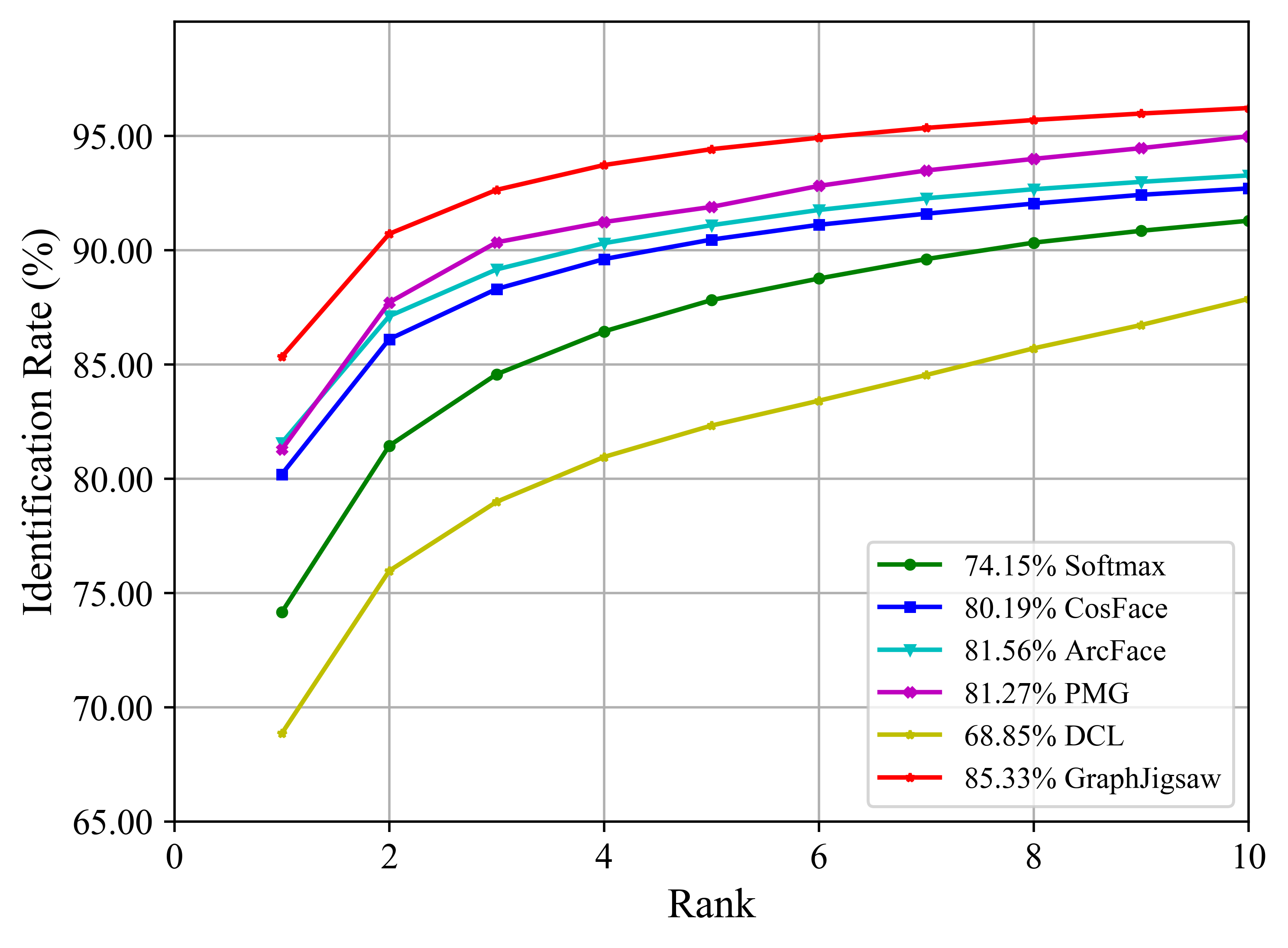}
		\label{fig:densenet-iqiyi}
    } 
	
    \subfloat[CMC curves on the \textbf{Danbooru} dataset (ResNet-50)]
    {
    \includegraphics[width=0.45\linewidth]{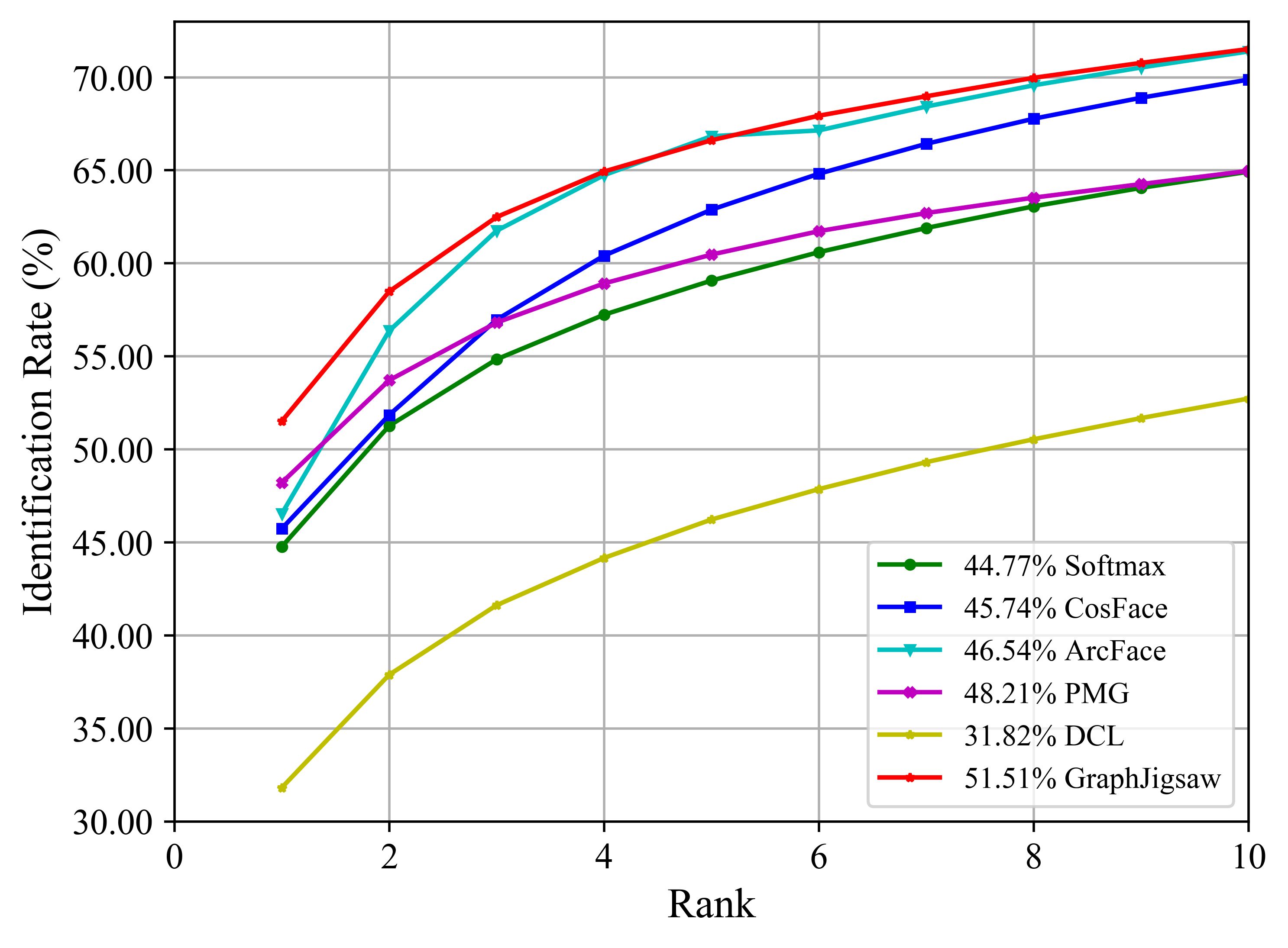}
		\label{fig:resnet-danbroo}
    }\hfil  
    \subfloat[CMC curves on the \textbf{Danbooru} dataset (DenseNet-169)]
    {
    \includegraphics[width=0.45\linewidth]{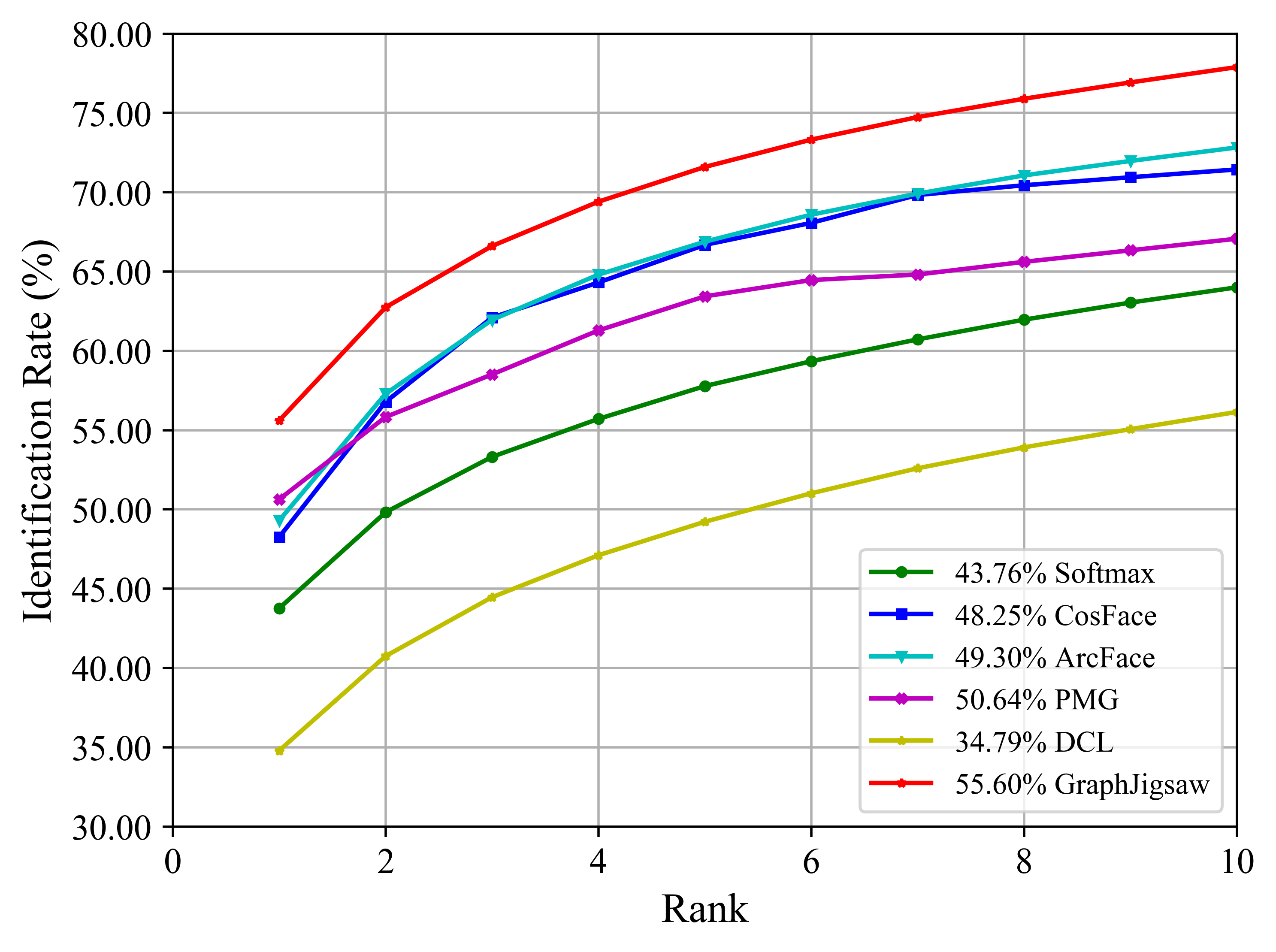}
		\label{fig:densenet-danbroo}
    }   

	\caption{Cumulative Match Characteristic (CMC) curves of the proposed GraphJigsaw and the other compared methods. The proposed GraphJigsaw shows a leading performance on various rank rates.}
	\label{fig:cmc_curve}
\end{figure*}



\textbf{Comparison with the FR methods:}  Tab.~\ref{tab:iqiyi_sate_of_the_art_resnet50},  \ref{tab:iqiyi_sate_of_the_art_densenet} report the experimental results on the iCartoonFace dataset.
The results in Tab.~\ref{tab:iqiyi_sate_of_the_art_resnet50} show GraphJigsaw outperforms the compared three FR methods (Softmax, CosFace, ArcFace) by 7.26\%, 3.38\%, 2.67\%  with the ResNet-50 backbone in Rank@1, Rank@5 and Rank@10, respectively. The similar improvements can be observed with the ResNet-101 backbone, as illustrated in Tab.~\ref{tab:iqiyi_sate_of_the_art_resnet50}. 
Tab.~\ref{tab:iqiyi_sate_of_the_art_densenet} shows the experimental results with a DenseNet-169 backbone on the iCartoonFace dataset.
Compared with the method in \cite{zheng2020cartoon} that jointly training the DenseNet-169 classification model with both the human and cartoon training images in a multi-task domain adaptation manner, our proposed GraphJigsaw obtains consistent improvements on  various rank rates. It is worth noting that the method in \cite{zheng2020cartoon} adopts CASIA-WebFace dataset \cite{yi2014learning} as the auxiliary training images, while GraphJigsaw merely exploits the cartoon face images in the iCartoonFace dataset.
Tab.~\ref{tab:iqiyi_sate_of_the_art_densenet} also show that our proposed GraphJigsaw outpperforms the compared FR methods by 3.77\%, 3.32\%, 2.96\% on the three rank rates with the DenseNet-169 network.  The experimental results in Tab.~\ref{tab:iqiyi_sate_of_the_art_resnet50},   \ref{tab:iqiyi_sate_of_the_art_densenet} show that improvements of our proposed GraphJigsaw are consistent across different network structures.

Tab.~\ref{tab:danboru_sate_of_the_art_resnet50},  \ref{tab:danbooru_sate_of_the_art_densenet} illustrate the experimental results on the Danbooru dataset.
It is clear that GraphJigsaw consistently outperforms the compared FR methods with consistent improvements with the ResNet-/DenseNet-based backbones.
On the two cartoon face datasets, the softmax-based FR method lags behind  GraphJigsaw because the standard CNN is biased towards texture, it is quite difficult to perceive the sparse and critical shape patterns of the cartoon faces.
CosFace and ArcFace obtain slight improvements over the softmax-based FR method. However, there is still a significant performance gap when compared with GraphJigsaw. 

\textbf{Comparison with the jigsaw-based methods:}  When comparing GraphJigsaw with DCL \cite{chen2019destruction} and PMG \cite{du2020fine} in table \ref{tab:iqiyi_sate_of_the_art_resnet50},  \ref{tab:danboru_sate_of_the_art_resnet50}, \ref{tab:iqiyi_sate_of_the_art_densenet} and \ref{tab:danbooru_sate_of_the_art_densenet}, we observe that our proposed GraphJigsaw always has a better performance, with higher performance on various rank rates. 
On the iCartoonFace dataset, GraphJigsaw outperforms the two jigsaw-based approaches with 20.33\%, 3.26\% improvements with the ResNet-50 backbone. 
GraphJigsaw also obtains 18.45\%, 2.92\% improvements with the ResNet-101 backbone and 16.48\%, 4.06\% with the DenseNet-169 backbone. 
The similar improvements can also be observed on the Danbooru dataset, as illustrated in Tab.~\ref{tab:danboru_sate_of_the_art_resnet50}, \ref{tab:danbooru_sate_of_the_art_densenet}.
The experimental results show GraphJigsaw-learned cartoon face representations are more discriminative compared with DCL \cite{chen2019destruction} and PMG \cite{du2020fine}. It is because directly training with the destructed images will introduce noisy visual patterns \cite{chen2019destruction}, and the features learned from these noise visual patterns are harmful to the classification task. 
Our proposed GraphJigsaw bypasses this dilemma by solving the jigsaw in the intermediate stages in the backbone network. In addition, the constructed jigsaw puzzles in the top-down stages of the classification model are solved progressively in a self-supervised manner, thus facilitates the model to spot the critical shape patterns gradually.

We visualize the Cumulative Match Characteristic (CMC) curves of the proposed GraphJigsaw and other compared methods based on the ResNet-50 and the DenseNet-169 backbones, as illustrated in Fig~\ref{fig:cmc_curve}.
The sub-figures (a), (b) show the CMC curves on the iCartoonFace dataset, and the sub-figures (c), (d) in Fig~\ref{fig:cmc_curve} illustrate the curves on the Danbooru dataset, respectively. The results in Fig~\ref{fig:cmc_curve} illustrate that GraphJigsaw obtains a leading performance among the state-of-the-art FR and the jigsaw-based methods, including the low-rank rate (Rank@1) and high-rank rate (Rank@10). 
The considerable improvements of GraphJigsaw over other compared methods can be observed in  Fig~\ref{fig:cmc_curve} (a),(b),(d).
This illustrates that the improvements of the proposed GraphJigsaw are consistent on various rank rates for cartoon face identification. We will show that GraphJigsaw is capable of spotting the critical shape characteristics of the cartoon faces with qualitative visualization results in Section \ref{sec:visualization}.

\begin{table}[htb]
	\centering
	\caption{Performance comparison of adding none, one or stage-wise GraphJigsaw in the ResNet-50 network. \textbf{Bold} denotes the best.}
	\label{tab:stage_wise_graphJigsaw}
	\begin{tabular}{c|c|c|c}
		\hline
		\multicolumn{4}{c}{Models trained on iCartoonFace dataset} \\
		\hline
		Method  & Rank@1 & Rank@5 & Rank@10  \\
		\hline
		w/o GraphJigsaw & 75.14 &  88.23 & 91.46 \\
		Stage 1  & 80.65 & 91.89 & 94.88 \\
		Stage 2   & 80.45 & 91.67 & 94.21 \\
		Stage 3  & 81.57 & 92.89 & 94.55 \\
		Stage 4  & 80.52 & 91.92 & 94.40 \\
		\textbf{Stage-wise} & \textbf{83.65} & \textbf{93.42} & \textbf{95.42} \\
		\hline
		\multicolumn{4}{c}{Models trained on Danbooru dataset} \\
		\hline
		w/o GraphJigsaw & 43.50 & 59.39 & 65.25 \\
		Stage 1  & 49.83 & 64.92 & 70.08 \\
		Stage 2   & 48.54 & 64.85 & 70.46 \\
		Stage 3  & 49.69 & 63.76 & 69.45 \\
		Stage 4  & 46.26 & 62.23 & 67.82 \\
		\textbf{Stage-wise} & \textbf{51.51} & \textbf{66.61} & \textbf{71.52} \\
		\hline
	\end{tabular}
\end{table}

\subsection{Ablation Study}
\label{sec:ablation_study}
We conducted comprehensive experiments to evaluate the influences of various hyper-parameters in GraphJigsaw in order to better understand our method. We also explored the influence of the progressive training strategy.

\begin{table}[htb]
	\centering
	\caption{Cartoon face recognition performance of GraphJigsaw with different number of graph vertices.   \textbf{Bold} denotes the best.}
	\label{tab:graph_vertices_ablation}
	\begin{tabular}{c|c|c|c}
		\hline
		\multicolumn{4}{c}{Models trained on the iCartoonFace dataset} \\
		\hline
		Method  & Rank@1 & Rank@5 & Rank@10  \\
		\hline
		$M$ = 2  & 81.66 & 91.96 & 94.11 \\
		\textbf{$\textbf{M}$ = 3}    & \textbf{83.65} & \textbf{93.42} & \textbf{95.42}\\
		$M$ = 4  & 82.68 & 92.85 & 95.12 \\
		$M$ = 5  & 82.07 & 92.64 & 95.00 \\
		\hline
		\multicolumn{4}{c}{Models trained on the Danbooru dataset} \\
		\hline
		$M$ = 2  & 49.24 & 65.19 & 70.35 \\
		\textbf{$\textbf{M}$ = 3}   & \textbf{52.74} & \textbf{67.40} & \textbf{72.15} \\
		$M$ = 4  & 49.32 & 64.71 & 69.99 \\
		$M$ = 5  & 50.82 & 64.10 & 69.43 \\
		\hline
	\end{tabular}
\end{table}

\begin{table}[htb]
	\centering
	\caption{Ablation studies on the iCartoonFace dataset. $\mathrm{T_{enc}}$ and $\mathrm{T_{dec}}$ denote the number of iterations in the \textit{face jigsaw encoding} and \textit{decoding} parts, respectively. \textbf{Bold} denotes the best.}
	\label{tab:iqiyi_different_input_output_resnet50}
	\begin{tabular}{c|c|c|c}
		\hline
		Method  & Rank@1 & Rank@5 & Rank@10  \\
		\hline
		$\mathbf{T_{enc}=1}$, $\mathbf{T_{dec}=1}$  & \textbf{83.65} & \textbf{93.42} & \textbf{95.42} \\
		$\mathrm{T_{enc}}$=1, $\mathrm{T_{dec}}$=2   & 80.53 & 91.79 & 94.33 \\
		$\mathrm{T_{enc}}$=2, $\mathrm{T_{dec}}$=1   & 78.97 & 90.79 & 93.40 \\
		
		\hline
		w/o jigsaw   & 77.98 & 90.23 & 93.23 \\
		\hline
		w/o progressive & 80.66 & 91.99 & 94.43\\
		\hline
	\end{tabular}
\end{table}

\begin{figure*}[htb!]
	\centering
	\includegraphics[width=0.8\textwidth]{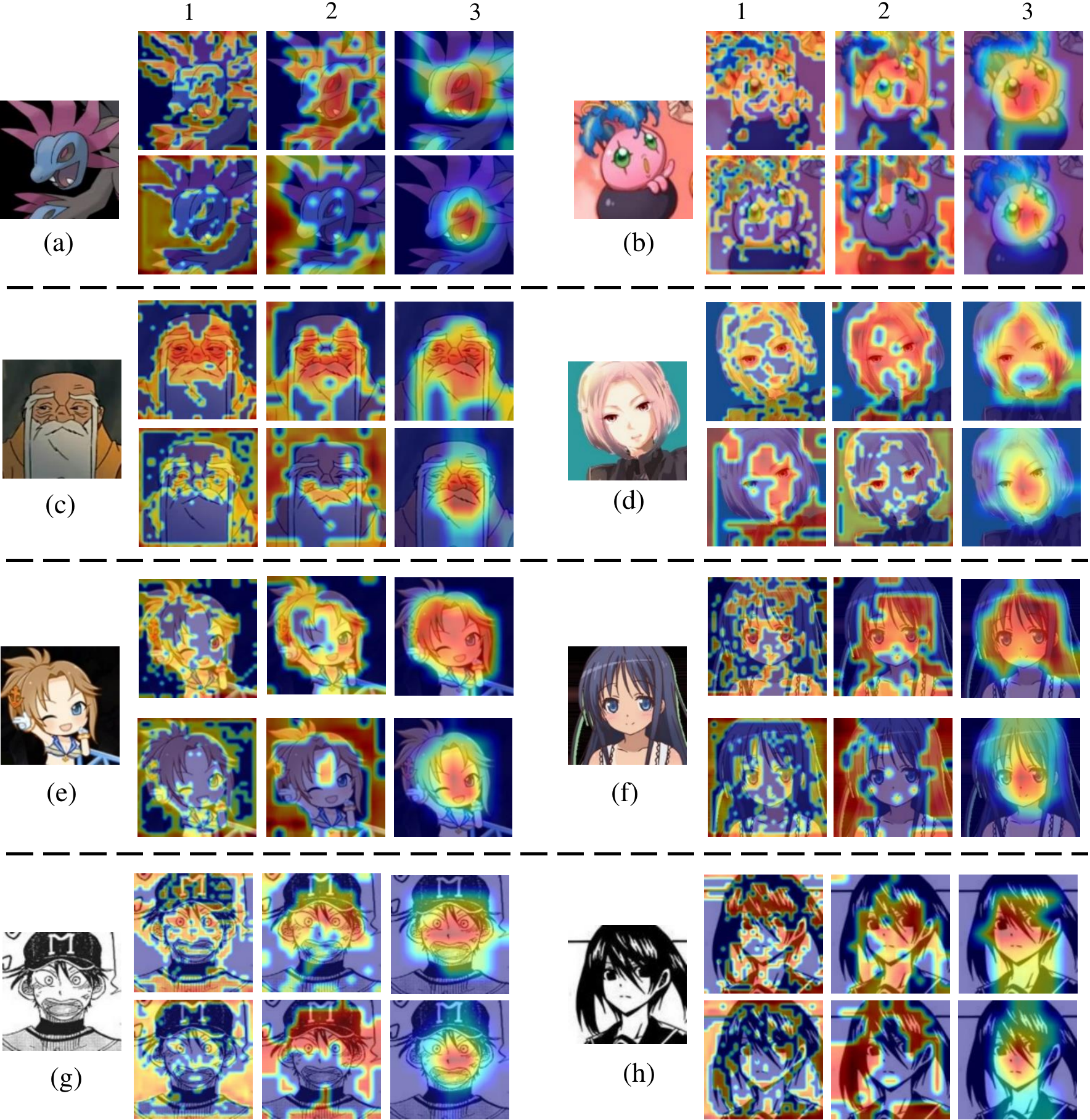}
	\caption{Attention maps on the iCartoonFace dataset with the ResNet-50 as the backbone network. For each sub-figure, the top and bottom row denote the visualization results of the last three stages' convolutional layers of our model and the baseline model, repspectively. Our proposed GraphJigsaw spots the critical shape characteristics gradually from column 1 to column 3 in each sub-figure. The baseline model hardly perceives the shape information in the early stages and only shows part correct attention in the last column. Better viewed in color and zoom in.}
	\label{fig:cartoon_visualization}
\end{figure*}

\begin{figure*}
		\centering
	\subfloat
	{
		\includegraphics[width=0.9\columnwidth]{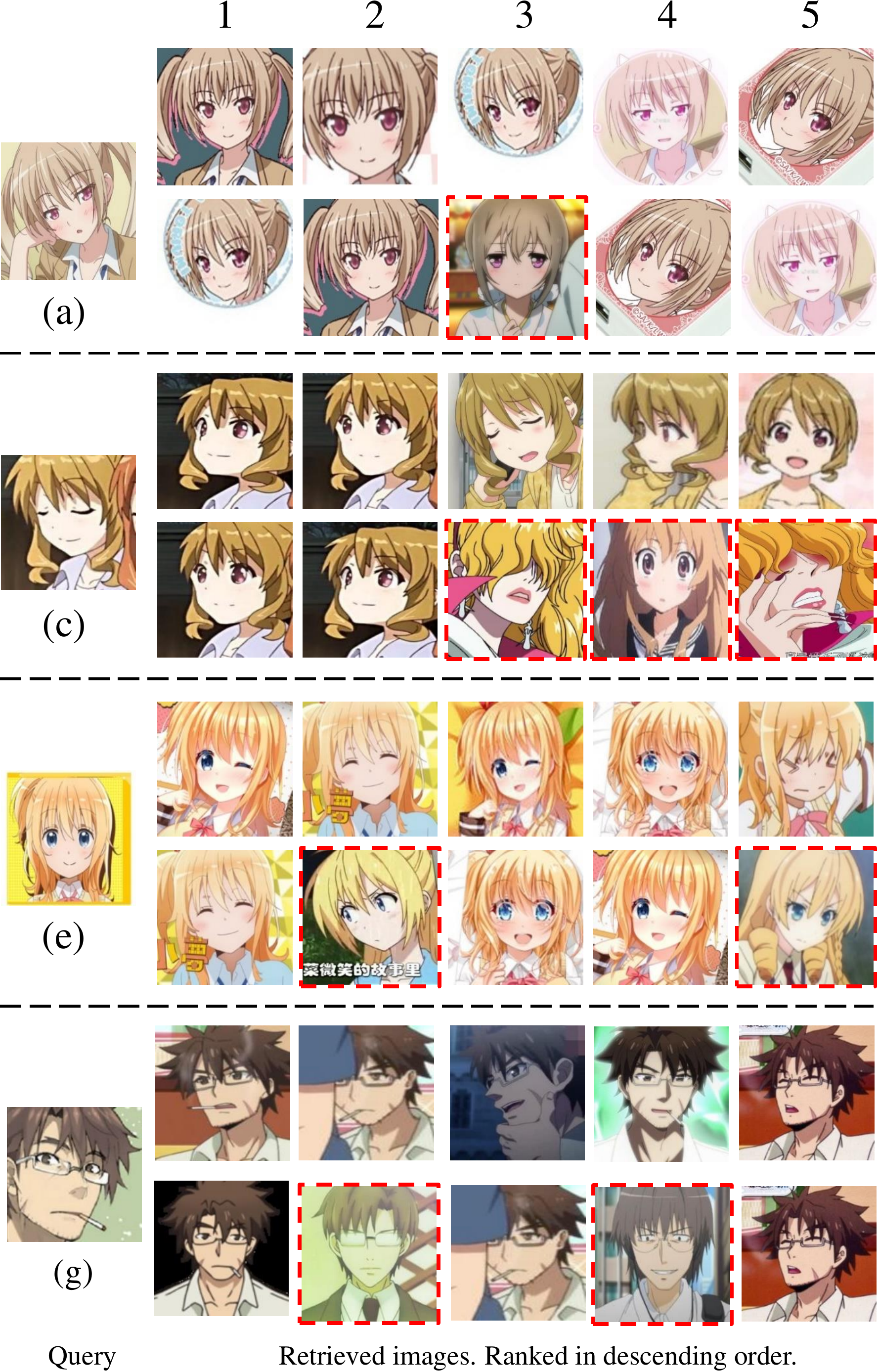}
		\label{fig:retrieval-a}
	}
	\subfloat
	{
		\includegraphics[width=0.9\columnwidth]{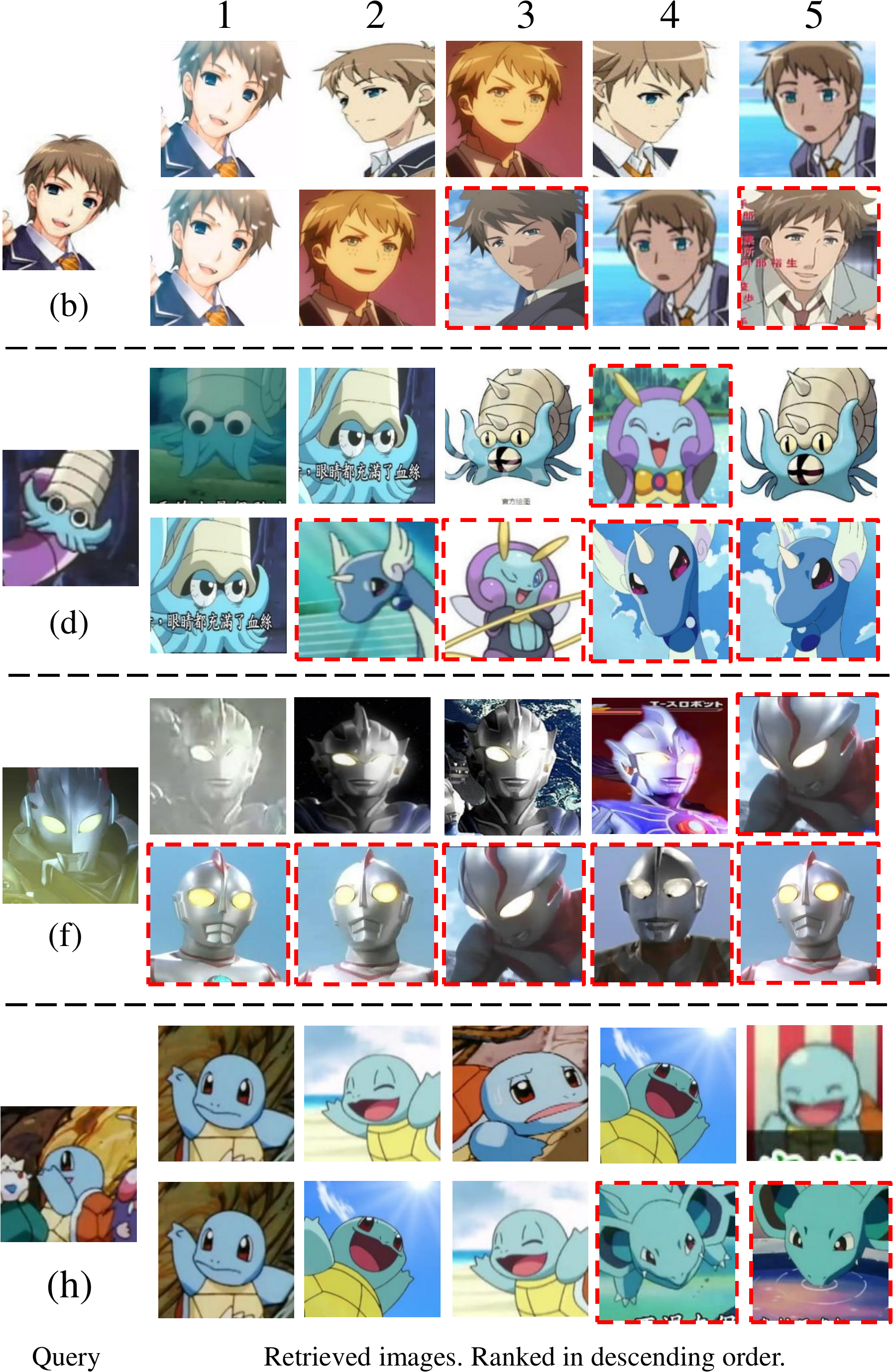}
		\label{fig:retrieval-b}
	}
	\caption{Retrieved images using the GraphJigsaw-learned features (top rows in (a-h)) and the features extracted from the baseline model (bottom rows in (a-h)). The images with red-dotted borders mean the incorrect retrieved images. The results indicate the proposed GraphJigsaw is capable of enhancing the intra-class compactness and inter-class discrepancy of the learned representations. Better viewed in color and zoom in.
	}
	\label{fig:retrieval}
\end{figure*} 

\textbf{Evaluation of the stage-wise GraphJigsaw:} 
We analyze the performance variations of our method by fully removing the jigsaw puzzle (w/o GraphJigsaw) or only adding one jigsaw puzzle during the training process.
As illustrated in Tab.~\ref{tab:stage_wise_graphJigsaw}, without the jigsaw puzzles, the classification model show a sharp deterioration in the cartoon face verification performance. Adding the jigsaw puzzle at arbitrary stage in the classification model yields the improvements over the baseline model. With the jigsaw puzzle at each stage in a top-down manner, our proposed method obtains the highest improvements. This suggests the stage-wise GraphJigsaw is reasonable and feasible.

\textbf{Evaluation of different number of graph vertices:} 
We analyze the influence of different graph vertices used in our proposed GraphJigsaw and the experimental results are illustrated in Tab.~\ref{tab:graph_vertices_ablation}. 
It is clear that our proposed GraphJigsaw obtains the best performance with $M=3$, which means that the jigsaw puzzle has $M^2 = 9$ fragments.
It suggests that the jigsaw puzzle with $M=2$ is easy to solve and the puzzle can be solved easily without spotting critical shape patterns of the input cartoon faces.
When it comes to $M=4$ or $M=5$, the jigsaw puzzle might be too confusing. It is because the cartoon faces usually have smooth and sparse color blocks,  the jigsaw fragments may contain little or even no shape information that can be exploited to solve the puzzle.

We additionally explore the best number of iterations in the face jigsaw encoding (Section \ref{sec:face_jigsaw_encoding})  and decoding (Section \ref{sec:face_jigsaw_decoding}) parts on the iCartoonFace dataset. The experimental results in Tab. \ref{tab:iqiyi_different_input_output_resnet50} show that GraphJigsaw obtains the leading performance with one iteration in the encoding and decoding part, respectively. This is in line with the conclusion in \cite{rong2019dropedge, zhou2020graph} that stacking more layers into a GCN causes over smoothing, eventually leading to features of graph vertices converging to the same value.

We show the cartoon face identification results without jigsaw (w/o jigsaw) or without the progressive training strategy (w/o progressive) in Tab.~\ref{tab:iqiyi_different_input_output_resnet50}. It is clear that the jigsaw task plays a vital role in spotting critical shape characteristics of input cartoon faces. Without solving the jigsaw task, the classification model shows a significant performance decrease. It is because without solving the jigsaw puzzle, the standard CNN-based classification model is biased towards recognizing textures rather than the shapes patterns \cite{baker2018deep}.
With the proposed progressive training strategy, the classification model is capable of solving the stage-wise jigsaw puzzle in a progressive manner, and absorbing the discriminative shape characteristics in a top-down manner.


\subsection{Visualization}
\label{sec:visualization}
\textbf{Visualization of the attention maps:} To investigate how well the shape characteristics are spotted progressively, we visualize the stage-wise attention maps of our method with Grad-CAM \cite{selvaraju2017grad} in Fig.~\ref{fig:cartoon_visualization}. We merely illustrate the activation maps of the last three stages as the map of the first stage is not visually informative. In each sub-figure in Fig.~\ref{fig:cartoon_visualization}, the top row show the attention maps of our proposed GraphJigsaw, the bottom row illustrate the attention maps of the baseline model that only exploits softmax loss for training.

As illustrated in Fig.~\ref{fig:cartoon_visualization}, our model shows more meaningful concentration on the input cartoon faces in a progressive manner in the top row in each sub-figure (a-h). In detail, our model focuses on various meaningful local regions at the second stage like hair, eyes, mouth.  When it comes to the third stage, our model starts to pay attention to the global shape of the input cartoon face images. Finally, the model spots the vital shape information that encodes the identity characteristics in the fourth stage. 
In Fig.~\ref{fig:cartoon_visualization} (a-f),  the input cartoon characters are distinguished by the shape characteristics, e.g., the triangular stips around the neck in (a), the ornaments on the hair in (b), the beard in (c), the hairstyle in (d), (e) (f). Our proposed method spots these critical characteristics gradually from column 1 to column 3 in each sub-figure.
As a comparison, the bottom rows in the sub-figures in Fig.~\ref{fig:cartoon_visualization} illustrate the baseline model hardly spots the critical areas with sparse shape characteristics and fails to perceive the shape patterns in the intermediate stages. The baseline model only shows the part correct attention at the last stage as shown in Fig.~\ref{fig:cartoon_visualization}.

We additionally illustrate two cartoon sketches in Fig.~\ref{fig:cartoon_visualization} (g) and (h), where the input cartoon faces are represented with binary-colored sketches and the identity information heavily rely on the shape rather than texture information. Our model successfully captures the shape patterns in such extreme conditions. The results also indicate that the baseline model is texture-biased and is not capable of spotting critical shape information without extra constraints. 

\textbf{Visualization of the image retrieval results:}  To further investigate the distinctiveness of the GraphJigsaw-learned features, we visualize the retrieved results of several query images using the GraphJigsaw-learned features and the baseline-learned features in Fig.~\ref{fig:retrieval}. The former and the latter are illustrated in the top and bottom row in each sub-figure  (a-h) in Fig.~\ref{fig:retrieval}. The images with red-dotted borders mean the incorrect retrieved images. When using the GraphJigsaw-learned features, the retrieved images show consistent cartoon face categories with the query images. For example, our model works well for the query images in sub-figures (a), (b), (c), (e), (f), (g) in Fig.~\ref{fig:retrieval} that are quite confusing as they contain little texture information.
The baseline model shows dramatically reduced performance and fails to capture the subtle shape difference between the query image and the mis-retrieved images. 
Besides, the sub-figures (g), (h) in Fig.~\ref{fig:retrieval} show the retrived results given the elves images. The proposed GraphJigsaw show consistent retrieval performance, while the baseline model fails in most cases. It indicates the GraphJigsaw is capable of enhancing the intra-class compactness and inter-class discrepancy of the learned representations.
The qualitative image retrieval results are in line with the quantitative experimental results in Section \ref{sec:comparison_stage_of_the_art}. They also highlight the importance of spotting the shape characteristics for cartoon face recognition.

\section{Conclusions}
Within this work we have presented GraphJigsaw to spot the critical shape patterns that are discriminative for cartoon face recognition. Our method is inspired by the fact that cartoon faces usually contain smooth color blocks and the key to recognize cartoon faces is to precisely spot their critical shape information.
We achieve this goal by constructing jigsaw tasks at various stages in the classification network and solving the tasks with the graph convolutional network in a progressive manner.  Extensive experiments demonstrated that our proposed method outperforms the state-of-the-art face recognition and the the jigsaw-based learning methods. Our method is conceptually elegant and we hope it will shed light on understanding and improving the performance of  cartoon face recognition.

\bibliographystyle{IEEEtran}
\bibliography{FOODR_mm2019}

\begin{IEEEbiography}[{\includegraphics[width=1in,height=1.25in,clip,keepaspectratio]{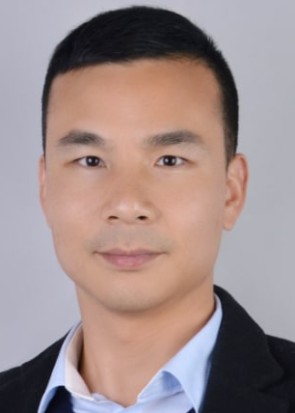}}]{Yong Li}  received the Ph.D. degree from Institute of Computing Technology (ICT), Chinese Academy of Sciences in 2020. He worked as a software engineer in Baidu company from 2015 to 2016. He has been a assistant professor at School of Computer Science and Engineering, Nanjing University of Science and Technology since 2020. His research interests include deep learning, self-supervised learning and affective computing.
\end{IEEEbiography}

\begin{IEEEbiography}[{\includegraphics[width=1in,height=1.25in,clip,keepaspectratio]{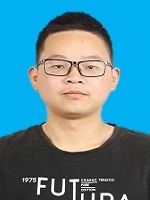}}]{LingJie Lao} received the B.S. degree from Ningbo University, Ningbo, China in 2020. He is currently working towards the M.S. degree in computer science and technology. His research interests include computer vision, face recognition and self-supervised learning.
\end{IEEEbiography}

\begin{IEEEbiography}[{\includegraphics[width=1in,height=1.25in,clip,keepaspectratio]{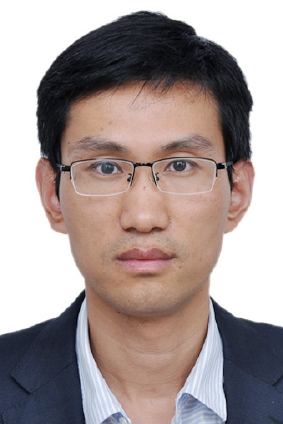}}]{Zhen Cui} received the Ph.D. degree from Institute of Computing Technology (ICT), Chinese Academy of Sciences in 2014. He was a Research Fellow in the Department of Electrical and Computer Engineering at National University of Singapore (NUS) from Sep 2014 to Nov 2015. He also spent half a year as a Research Assistant on Nanyang Technological University (NTU) from Jun 2012 to Dec 2012. Currently, he is a Professor of Nanjing University of Science and Technology, China.
	His research interests cover computer vision, pattern recognition and machine learning, especially focusing on  vision perception and computation, graph deep learning, etc.
\end{IEEEbiography}

\begin{IEEEbiography}[{\includegraphics[width=1in,height=1.25in,clip,keepaspectratio]{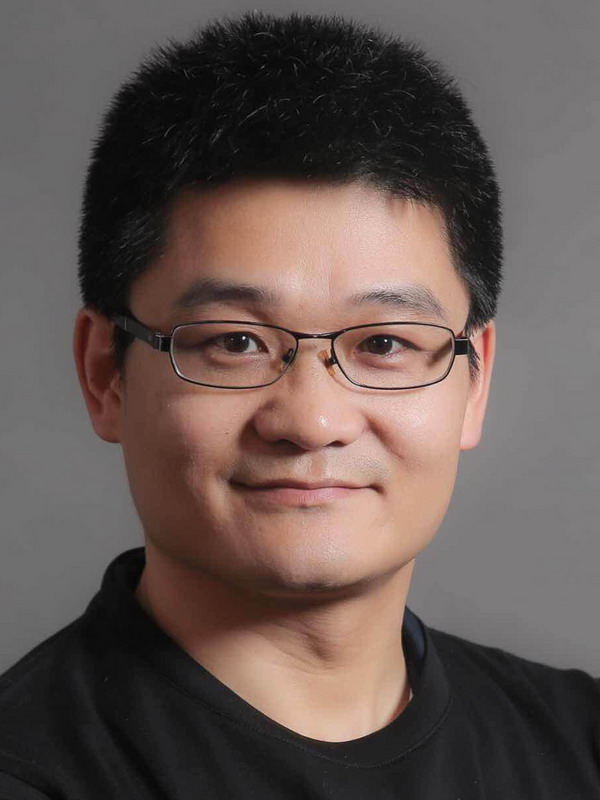}}]{Shiguang Shan} received M.S. degree in computer science from the Harbin Institute of Technology, Harbin, China, in 1999, and Ph.D. degree in computer science from the Institute of Computing Technology (ICT), Chinese Academy of Sciences (CAS), Beijing, China, in 2004. He joined ICT, CAS in 2002 and has been a Professor since 2010. He is now the deputy director of the Key Lab of Intelligent Information Processing of CAS. His research interests cover computer vision, pattern recognition, and machine learning. He especially focuses on face recognition related research topics. He has published more than 200 papers in refereed journals and proceedings in the areas of computer vision and pattern recognition. He has served as Area Chair for many international conferences including ICCV'11, ICPR'12, ACCV'12, FG'13, ICPR'14, ICASSP'14, ACCV'16, ACCV18, FG'18, and BTAS'18. He is Associate Editors of several international journals including IEEE Trans. on Image Processing, Computer Vision and Image Understanding, Neurocomputing, and Pattern Recognition Letters. He is a recipient of the China's State Natural Science Award in 2015, and the China’s State S\&T Progress Award in 2005 for his research work. He is also personally interested in brain science, cognitive neuroscience, as well as their interdisciplinary researche topics with AI.
\end{IEEEbiography}

\begin{IEEEbiography}[{\includegraphics[width=1in,height=1.25in,clip,keepaspectratio]{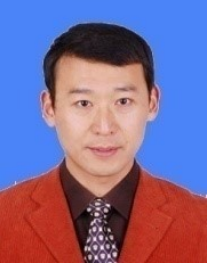}}]{Jian Yang}
	received the PhD degree from Nanjing University of Science and Technology (NUST), on the subject of pattern recognition and intelligence systems in 2002. In 2003, he was a postdoctoral researcher at the University of Zaragoza. From 2004 to 2006, he was a Postdoctoral Fellow at Biometrics Centre of Hong Kong Polytechnic University. From 2006 to 2007, he was a Postdoctoral Fellow at Department of Computer Science of New Jersey Institute of Technology. Now, he is a Chang-Jiang professor in the School of Computer Science and Technology of NUST. He is the author of more than 100 scientific papers in pattern recognition and computer vision. His journal papers have been cited more than 4000 times in the ISI Web of Science, and 9000 times in the Web of Scholar Google. His research interests include pattern recognition, computer vision and machine learning. Currently, he is/was an associate editor of Pattern Recognition Letters, IEEE Trans. Neural Networks and Learning Systems, and Neurocomputing. He is a Fellow of IAPR.
\end{IEEEbiography}







\end{document}